\documentclass[10pt,twocolumn,letterpaper]{article}

\usepackage{cvpr}              

\usepackage[accsupp]{axessibility} 

\usepackage{graphicx}
\usepackage{amsmath}
\usepackage{amssymb}
\usepackage{booktabs}
\usepackage{bm}
\usepackage{enumitem}

\usepackage[ruled]{algorithm2e}

\usepackage[pagebackref,breaklinks,colorlinks]{hyperref}

\usepackage[capitalize]{cleveref}
\crefname{section}{Sec.}{Secs.}
\Crefname{section}{Section}{Sections}
\Crefname{table}{Table}{Tables}
\crefname{table}{Tab.}{Tabs.}

\def\cvprPaperID{8015} 
\def\confName{CVPR}
\def\confYear{2023}

\begin{document}

\title{LayoutDM: Transformer-based Diffusion Model for Layout Generation}
\author{Shang Chai, Liansheng Zhuang\thanks{Corresponding author.}\\
University of Science and Technology of China\\
{\tt\small chaishang@mail.ustc.edu.cn, lszhuang@ustc.edu.cn}
\and
Fengying Yan\\
Tianjin University\\
{\tt\small fengying@tju.edu.cn}
}

\maketitle

\setlength{\textfloatsep}{0.25cm}
\setlength{\floatsep}{0.15cm}

\begin{abstract}
Automatic layout generation that can synthesize high-quality layouts is an important tool for graphic design in many applications. Though existing methods based on generative models such as Generative Adversarial Networks (GANs) and Variational Auto-Encoders (VAEs) have progressed, they still leave much room for improving the quality and diversity of the results. 
Inspired by the recent success of diffusion models in generating high-quality images, this paper explores their potential for conditional layout generation and proposes Transformer-based Layout Diffusion Model (LayoutDM) by instantiating the conditional denoising diffusion probabilistic model (DDPM) with a purely transformer-based architecture. 
Instead of using convolutional neural networks, a transformer-based conditional Layout Denoiser is proposed to learn the reverse diffusion process to generate samples from noised layout data. Benefitting from both transformer and DDPM, our LayoutDM is of desired properties such as high-quality generation, strong sample diversity, faithful distribution coverage, and stationary training in comparison to GANs and VAEs. Quantitative and qualitative experimental results show that our method outperforms state-of-the-art generative models in terms of quality and diversity. 
\end{abstract}

\section{Introduction}
\label{sec:intro}
Layouts, \emph{i.e.} the arrangement of the elements to be displayed in a design, play a critical role in many applications from magazine pages to advertising posters to application interfaces. A good layout guides viewers’ reading order and draws their attention to important information. 
The semantic relationships of elements, the reading order, canvas space allocation and aesthetic principles must be carefully decided in the layout design process.
However, manually arranging design elements to meet aesthetic goals and user-specified constraints is time-consuming. To aid the design of graphic layouts, the task of layout generation aims to generate design layouts given a set of design components with user-specified attributes. Though meaningful attempts are made~\cite{10.1145/3306346.3322971, 10.1007/978-3-030-58580-8_29, 10.1145/3411764.3445117, Patil2020READRA, 9710883, Kikuchi2021, Kong2021BLTBL, 8948239, 9106863, 9711031, Coarse2fine, wang2021aesthetic, 9578374}, it is still challenging to generate realistic and complex layouts, because many factors need to be taken into consideration, such as design elements, their attributes, and their relationships to other elements. 

Over the past few years, generative models such as Generative Adversarial Networks (GANs)~\cite{goodfellow2014generative} and Variational Auto-Encoders (VAEs)~\cite{Kingma2014AutoEncodingVB} have gained much attention in layout generation, as they have shown a great promise in terms of faithfully learning a given data distribution and sampling from it. GANs model the sampling procedure of a complex distribution that is learned in an adversarial manner, while VAEs seek to learn a model that assigns a high likelihood to the observed data samples. 
Though having shown impressive success in generating high-quality layouts, these models have some limitations of their own. 
GANs are known for potentially unstable training and less distribution coverage due to their adversarial training nature~\cite{Brock2019LargeSG, Miyato2018SpectralNF, Brock2017NeuralPE}, so they are inferior to state-of-the-art likelihood-based models (such as VAEs) in terms of diversity~\cite{razavi2019generating, nichol2021improved, Nash2021GeneratingIW}. 
VAEs can capture more diversity and are typically easier to scale and train than GANs, but still fall short in terms of visual sample quality and sampling efficiency~\cite{larsen2016autoencoding}.

Recently, diffusion models such as denoising diffusion probabilistic model (DDPM)~\cite{NEURIPS2020_4c5bcfec} have emerged as a powerful class of generative models, capable of producing high-quality images comparable to those of GANs. Importantly, they additionally offer desirable properties such as strong sample diversity, faithful distribution coverage, a stationary training objective, and easy scalability. This implies that diffusion models are well suited for learning models of complex and diverse data, which also motivates us to explore the potential of diffusion-based generative models for graphic layout generation.

Though diffusion models have shown splendid performance in high-fidelity image generation~\cite{NEURIPS2021_49ad23d1, NEURIPS2020_4c5bcfec, rombach2021highresolution, 10.5555/3045118.3045358, song2019generative}, it is still a sparsely explored area and provides unique challenges to develop diffusion-based generative models for layout generation. 
First, diffusion models often use convolutional neural networks such as U-Net~\cite{10.1007/978-3-319-24574-4_28} to learn the reverse process to construct desired data samples from the noise. However, a layout is a non-sequential data structure consisting of varying length samples with discrete (classes) and continuous (coordinates) elements simultaneously, instead of pixels laid on a regular lattice. Obviously, convolutional neural networks are not suitable for layout denoising, which prevents diffusion models from being directly applied to layout generation. 
Second, the placement and sizing of a given element depend not only on its attributes (such as category label) but also on its relationship to other elements. How to incorporate the attributes knowledge and model the elements' relationship in diffusion models is still an open problem. 
Since diffusion models are general frameworks, they leave room for adapting the underlying neural architectures to exploit the properties of the data.

Inspired by the above insights, by instantiating the conditional denoising diffusion probabilistic model (DDPM) with a transformer architecture, this paper proposes Transformer-based Layout Diffusion Model (\emph{i.e.}, LayoutDM) for conditional layout generation given a set of elements with user-specified attributes. The key idea is to use a purely transformer-based architecture instead of the commonly used convolutional neural networks to learn the reverse diffusion process from noised layout data. 
Benefitting from the self-attention mechanism in transformer layers, LayoutDM can efficiently capture high-level relationship information between elements, and predict the noise at each time step from the noised layout data. Moreover, the attention mechanism also helps model another aspect of the data - namely a varying and large number of elements. 
Finally, to generate layouts with desired attributes, LayoutDM designs a conditional Layout Denoiser (cLayoutDenoiser) based on a transformer architecture to learn the reverse diffusion process conditioned on the input attributes. Different from previous transformer models in the context of NLP or video, cLayoutDenoiser omits the positional encoding which indicates the element order in the sequence, as we do not consider the order of designed elements on a canvas in our setting. 
In comparison with current layout generation approaches (such as GANs and VAEs), our LayoutDM offers several desired properties such as high-quality generation, better diversity, faithful distribution coverage, a stationary training objective, and easy scalability. 
Extensive experiments on five public datasets show that LayoutDM outperforms state-of-the-art methods in different tasks.


In summary, our main contributions are as follows:
{
\begin{itemize}[itemsep=0.1ex, parsep=0.7ex, leftmargin=0.6cm]
    \item This paper proposes a novel LayoutDM to generate high-quality design layouts for a set of elements with user-specified attributes. Compared with existing methods, LayoutDM is of desired properties such as high-quality generation, better diversity, faithful distribution coverage, and stationary training. To our best knowledge, LayoutDM is the first attempt to explore the potential of diffusion model for graphic layout generation. 
    \item  This paper explores a new class of diffusion models by replacing the commonly-used U-Net backbone with a transformer, and designs a novel cLayoutDenoiser to reverse the diffusion process from noised layout data and better capture the relationship of elements.
    
    \item Extensive experiments demonstrate that our method outperforms state-of-the-art models in terms of visual perceptual quality and diversity on five diverse layout datasets.
\end{itemize}
}

\begin{figure*}[t]
\centering
	\subcaptionbox{LayoutDM architecture\label{fig:a}}{\includegraphics[width = 0.48\textwidth]{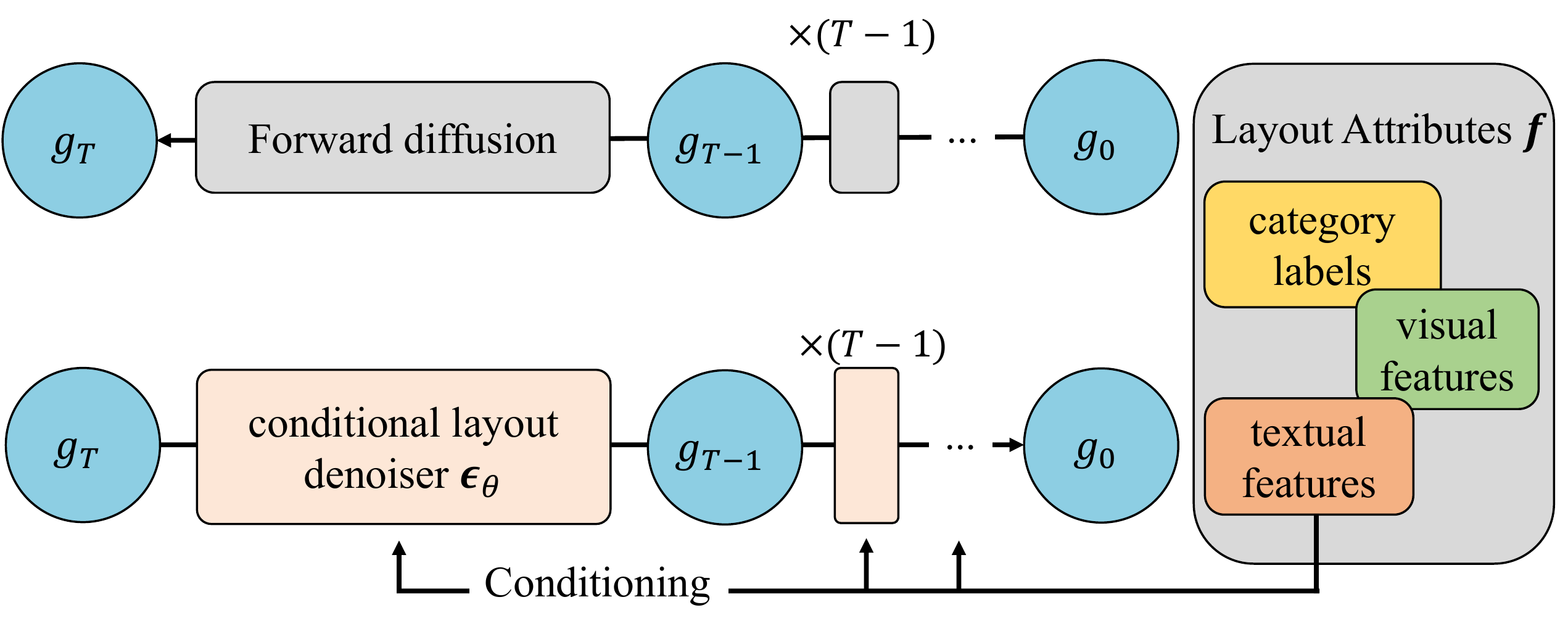}}
	\subcaptionbox{Architecture of conditional layout denoiser $\bm\epsilon_\theta$\label{fig:b}}{\includegraphics[width = 0.515\textwidth]{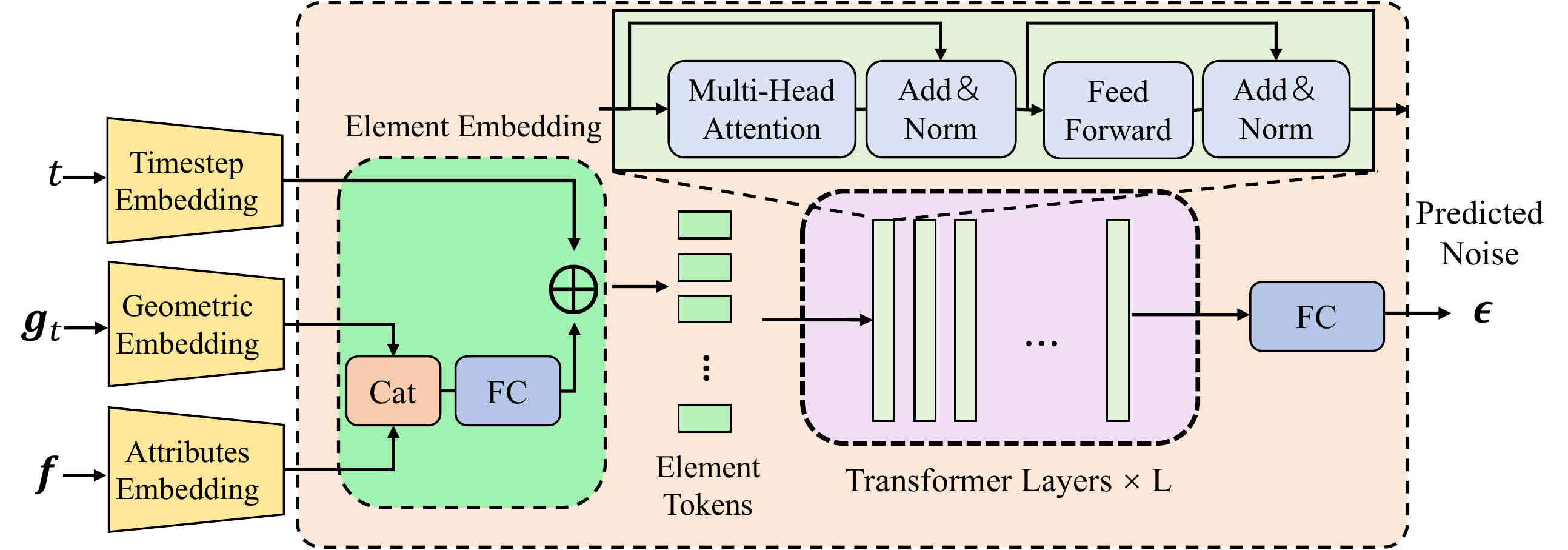}}
\caption{a) Architecture of LayoutDM. It consists of a forward diffusion process and a reverse process modeled by a conditional layout denoiser $\bm\epsilon_\theta$. 
   b) Architecture of our transformer-based conditional layout denoiser, cLayoutDenoiser. cLayoutDenoiser predicts the added noise conditioned on the layout attributes $\bm{f}$ and time step $t$.}
\label{fig:label}
\vspace{-0.4cm}
\end{figure*}

\section{Related Work}
\label{sec:related work}

\subsection{Layout Generation}

Automatic layout generation has been widely studied for a long time. 
Early approaches to layout generation~\cite{10.1145/2702123.2702149,6777138} embed design rules into manually-defined energy functions.
In recent years, generative model based methods are increasingly progressed. LayoutGAN~\cite{8948239} and LayoutVAE~\cite{9010959} are the first attempts to utilize GAN and VAE to generate graphic and scene layouts. NDN~\cite{10.1007/978-3-030-58580-8_29} represents the relative positional relationship of elements as a graph and uses a graph neural network based conditional VAE to generate graphic layouts. READ~\cite{Patil2020READRA} uses heuristics to determine the relationships between elements and trains a Recursive Neural Network (RNN)~\cite{10.5555/3104482.3104499,socher-etal-2013-recursive} based VAE to learn the layout distribution. CanvasVAE~\cite{9711031} generates vector graphic documents which contain structured information about canvas and elements. VTN~\cite{9578374} and Coarse2fine~\cite{Jiang2022CoarsetoFineGM} deploy self-attention based VAEs to generate graphic layouts, making progress in diversity and perceptual quality. LayoutTransformer~\cite{9710883} and BLT~\cite{Kong2021BLTBL} define layouts as discrete sequences and exploit the efficiency of transformer and bidirectional-transformer in structured sequence generation. LayoutNet~\cite{10.1145/3306346.3322971}, TextLogo3K~\cite{wang2021aesthetic} and ICVT~\cite{10.1145/3503161.3548332} propose conditional layout generative models which can utilize additional attributes about design elements or entire layouts to aid layout generation in different application scenarios. LayoutGAN++~\cite{Kikuchi2021} designs a transformer-based generator and discriminator and generates graphic layouts conditioned on the given element category labels.

\subsection{Diffusion Model}
Diffusion models~\cite{NEURIPS2020_4c5bcfec,song2020denoising} have proved its capability of generating high-quality and diverse samples~\cite{rombach2021highresolution, NEURIPS2021_49ad23d1, NEURIPS2020_4c5bcfec} and have recently achieved state-of-the-art results on several benchmark generation tasks~\cite{NEURIPS2021_49ad23d1}. The diffusion model uses diffusion processes to model the generation and defines the sampling of data as the process of gradually denoising from complete Gaussian noise. The forward process gradually adds Gaussian noise to the data from a predefined noise schedule
until time step $T$. The reverse process uses a neural backbone often implemented as a U-Net~\cite{NEURIPS2021_49ad23d1, NEURIPS2020_4c5bcfec, 10.1007/978-3-319-24574-4_28, song2020score} to parameterize the conditional distribution $p(\bm{x}_{t-1}|\bm{x}_t)$. In this work, we instantiate a conditional diffusion model to achieve conditional layout generation.

\section{Our Method}
\label{sec:method}


\subsection{Layout Representation}
In our model, each layout consists of a set of elements, and each element is described by both geometric parameters (\ie location and size) and its attribute (\eg category label or textual features). 
Formally, a layout $\bm{l}$ is denoted as a flattened sequence of integer indices:
$$\bm{l}=(g_1,f_1,g_2,f_2,\cdots,g_i,f_i,\cdots,g_N,f_N),$$
where $N$ is the number of elements in the layout. $g_i=[x_i, y_i, w_i, h_i]$ is a vector that presents the geometric parameters (center coordinates and size) of the $i$-th element in the layout. $f_i$ is the attributes of $i$-th element which might be category label or textual features. 
For the sake of convenience, the sequence $\bm{g} = (g_1, g_2, \cdots, g_i \cdots, g_N)$ is named layout geometric parameters, the sequence $\bm{f}=(f_1, f_2, \cdots, f_i \cdots, f_N)$ is named layout attributes.  
Note here that, the elements in a layout are unordered, so swapping the items in sequence $\bm{g}$ and $\bm{f}$ does not affect the meaning of the sequences. We normalize the geometric parameters, \ie, $[x_i, y_i, w_i, h_i],\ i=1,2,\cdots,N$, to the interval $[-1, 1]$
. In this way, layouts have a uniform structured representation.

\subsection{The LayoutDM Architecture}
\label{sec:4.2}

\cref{fig:a} illustrates the architecture of LayoutDM. 
From a high-level perspective, our LayoutDM is an instance of conditional denoising diffusion probabilistic model (DDPM) with a transformer architecture suitable for layout data. DDPM learns to model the Markov transition from simple distribution to layout data distribution and generates diverse samples through sequential stochastic transitions. To generate desired layouts, LayoutDM uses the input attributes to guide the generative process in DDPM. We follow the method described in Classifier-Free Diffusion Guidance~\cite{Ho2022ClassifierFreeDG} to realize conditional DDPM, and set the guidance strength $w$ to zero for simplicity.

Specifically, let $q(\bm{g}_0|\bm{f})$ be the unknown conditional data distribution, where $\bm{g}_0$ is the geometric parameters of real layouts, and $\bm{f}$ is the layout attributes. LayoutDM models the conditional distribution $q(\bm{g}_0|\bm{f})$ by two processes: a forward diffusion process and a reverse denoising diffusion process. First, LayoutDM defines the forward diffusion process $q(\bm{g}_t|\bm{g}_{t-1})$ which maps layout data to noise by gradually adding Gaussian noise at each time step $t$:
\begin{equation}
    q(\bm{g}_t|\bm{g}_{t-1})=\mathcal{N}(\bm{g}_t;\sqrt{1-\beta_t}\bm{g}_{t-1},\beta_t\mathbf{I})
\end{equation}
where $\{\beta_t\}_{t=1}^T$  are forward process variances. 

Then, LayoutDM defines the conditional reverse diffusion process $p(\bm{g}_{t-1}|\bm{g}_t,\bm{f})$ which performs iterative denoising from pure Gaussian noise to generate high-quality layouts conditioned on layout attributes $\bm{f}$:
\begin{equation}
    p_\theta(\bm{g}_{t-1}|\bm{g}_t,\bm{f}) = \mathcal{N}(\bm{g}_{t-1};\bm\mu_\theta(\bm{g}_t, t, \bm{f}), \sigma_t^2\mathbf{I})
\end{equation}
where $\sigma_t$ is the constant variance following ~\cite{NEURIPS2020_4c5bcfec}, $\bm\mu_\theta$ is the mean of the Gaussian distribution computed by a neural network, and $\theta$ is the parameters of the network. 
As shown in Ho \etal~\cite{NEURIPS2020_4c5bcfec}, we can reparameterize the mean to make the neural network learn the added noise at time step $t$ instead. 
In this way, $\bm{\mu}_\theta$ can be reparameterized as follows:
\begin{equation}
    \bm\mu_\theta(\bm{g}_t, t, \bm{f})=\frac{1}{\sqrt{\alpha_t}}(\bm{g}_t-\frac{\beta_t}{\sqrt{1-\overline{\alpha}_t}}\bm\epsilon_\theta(\bm{g}_t, t, \bm{f}))
\end{equation}
where $t$ is the time step, $\{\beta_t\}_{t=1}^T$ are forward process variances, $\alpha_t=1-\beta_t$, and  $\overline{\alpha}_t=\prod_{s-1}^t\alpha_s$. $\bm\epsilon_\theta(\bm{g}_t, t, \bm{f})$ is the neural network to predict the added noise for layout geometric parameters conditioned on elements' attributes at time step $t$. We also call the neural network $\bm\epsilon_\theta(\bm{g}_t, t, \bm{f})$ as \emph{conditional layout denoiser} (cLayoutDenoiser).



\subsection{Conditional Layout Denoiser}

The inputs to cLayoutDenoiser are layout geometric parameters $\bm{g}_t$, layout attributes $\bm{f}$ and time step $t$. 
To deal with sequences data, the conditional layout denoiser $\bm\epsilon_\theta(\bm{g}_t, t, \bm{f})$ employs a purely transformer-based architecture instead of convolutional neural networks as its backbone. The architecture of cLayoutDenoiser is illustrated in \cref{fig:b}. Benefitting from the transformer architecture, cLayoutDenoiser can deal with the sequence with various lengths, and capture the relationships among elements. Moreover, cLayoutDenoiser adds ``attributes embeddings" to the input ``geometric embeddings", so as to guide the reverse diffusion process at each time step $t$. Formally, the architecture of cLayoutDenoiser can be described as follow:
\begin{align}
    \bm{h}_f &= \mathrm{AttributesEmbedding}(\bm{f})\\
    \bm{h}_g &= \mathrm{GeometricEmbedding}(\bm{g}_t)\\
    \mathbf{E} &= \mathrm{ElementEmbedding}(\bm{h}_f,\bm{h}_g,\mathrm{TE}(t))\\
    \mathbf{E}' &= \mathrm{TransformerLayers}(\mathbf{E})\\
    \bm\epsilon &= \mathrm{FC}(\mathbf{E}')
\end{align}
where $\bm{f}$ is the layout attributes, $\bm{g}_t$ is the noised layout geometric parameters, $\bm{h}_f$ and $\bm{h}_g$ are their hidden representations. $\mathbf{E}$ is the element tokens computed by element embedding module, $\mathbf{E'}$ is intermediate feature. $\mathrm{TE}(t)$ denotes the timestep embedding, and $\bm\epsilon$ is the predicted noise. Note here that, since the element order in the sequence makes no sense in our setting, cLayoutDenoiser omits the positional encoding, which is different from existing transformers in the context of NLP or video.

\textbf{Geometric, attributes and timestep embedding.} Three embedding modules are used to learn meaningful representations for noised layout geometric parameters  $\bm{g}_t$, layout attributes $\bm{f}$ and time step $t$. 
Geometric embedding projects layout geometric parameters to a specific dimension, aiming to find a more efficient feature space than the original coordinate space.
Feature embedding learns the continuous features of discrete element attributes by embedding them into a specific dimension. 
Following~\cite{NEURIPS2020_4c5bcfec}, we condition cLayoutTransformer on time step $t$ by adding a sinusoidal time embedding $\mathrm{TE}(t)$ to make the network aware at which time step it is operating.

\textbf{Element embedding.} Element embedding module calculates the element tokens used as the input of transformer layers. Element tokens should contain geometric, attributes and time step information, so that transformer can efficiently capture the relationship information between elements conditioned on the time step $t$ and layout attributes $\bm{f}$. We concatenate the geometric embedding and attributes embedding, and then use a fully-connected layer to fuse element representation. We further perform an element-wise plus operation with timestep embedding on the results to finally obtain the element tokens.


\textbf{Transformer layers}. 
Generating an effective layout requires understanding the relationships between layout elements. Self-attention mechanism in Transformer~\cite{NIPS2017_3f5ee243} has proven effective in capturing high-level relationships between lots of elements in layout generation~\cite{9578374,Coarse2fine,9710883}. In this paper, we adopt multihead attention mechanism to capture relationship information between elements from element tokens.
We stack multiple transformer layers (8 in our model) to enable cLayoutDenoiser to capture relationships between layout elements from the element tokens. Positional encoding is omitted because of the unordered nature of elements. 
\begin{align}
    \hat{\mathbf{E}} &= \mathrm{LayerNorm}(\mathbf{E}^{l-1}+\mathbf{Head}(\bm{e}^{l-1}_1,\cdots,\bm{e}^{l-1}_N))\\
    \mathbf{E}^{l} &= \mathrm{LayerNorm}(\hat{\mathbf{E}} + \mathrm{FFN}(\hat{\mathbf{E}}))
\end{align}

\noindent where $l=1,\cdots,L$ denotes the layer index, $\mathbf{Head}$, $\mathrm{LayoutNorm}$ and $\mathrm{FFN}$ denote multi-head attention layer, Layer Normalization~\cite{Ba2016LayerN} and fully connected feed-forward network. $\mathbf{E}^{l-1}=(\bm{e}_1^{l-1},\cdots,\bm{e}_N^{l-1})$ are the intermediate element tokens used as the input of $l$-th transformer layer. 



\SetKwInput{KwIn}{Require}
\begin{algorithm}[t]
	\caption{Training LayoutDM}
	\label{alg:algorithm1}
	\KwIn{conditional layout denoiser $\bm\epsilon_\theta$}
	\Repeat{converged}{
	Sample $(\bm{g}_0, \bm{f})\sim q_{data}$\;
	$t\sim \mathrm{Uniform}(\{1,\cdots,T\})$\;
	$\bm\epsilon\sim \mathcal{N}(\mathbf{0}, \mathbf{I})$\;
	$\bm{g}_t=\sqrt{\overline{\alpha}}_t\bm{g}_0+\sqrt{1-\overline{\alpha}_t}\bm\epsilon$\;
	Take gradient descent step on\\
	\qquad $\nabla_\theta\|\bm\epsilon-\bm\epsilon_\theta(\bm{g}_t, t, \bm{f})\|^2$\;
	}
\end{algorithm}

\SetKwInput{KwIn}{Input}
\begin{algorithm}[t]
	\caption{Sampling}
	\label{alg:algorithm2}
	\KwIn{layout attributes $\bm{f}$}
	\KwOut{layout geometric parameters $\bm{g}_0$}
	$\bm{g}_T\sim \mathcal{N}(\mathbf{0}, \mathbf{I})$\;
	\For{$t=T,\cdots,1$}{
	$\mathbf{z}\sim \mathcal{N}(\mathbf{0}, \mathbf{I})$ if $t>1$, else $\mathbf{z}=\mathbf{0}$\;
	$\bm{g}_{t-1}=\frac{1}{\sqrt{\alpha_t}}(\bm{g}_t-\frac{1-\alpha_t}{\sqrt{1-\overline{\alpha}_t}}\bm\epsilon_\theta(\bm{g}_t,t,\bm{f}))+\sigma_t\mathbf{z}$
	}
	\Return{$\bm{g}_0$}
\end{algorithm}

\subsection{Training and Inference}

Following denoising diffusion probabilistic model~\cite{NEURIPS2020_4c5bcfec}, we optimize random terms $L_t$ which are the KL divergences between $p_\theta(\bm{g}_{t-1}|\bm{g}_t,\bm{f})$ and forward process posteriors. 
After simplifying the objective function following the method in ~\cite{NEURIPS2020_4c5bcfec}, the final loss function is as follows:
\begin{equation}
    \begin{aligned}
    L_{simple}(\theta) &= \|\bm\epsilon-\bm\epsilon_\theta(\bm{g}_t, t, \bm{f})\|^2\\
    &=\|\bm\epsilon-\bm\epsilon_\theta(\sqrt{\overline{\alpha}_t}\bm{g}_0+\sqrt{(1-\overline{\alpha}_t)}\bm\epsilon, t, \bm{f})\|^2
\end{aligned}
\end{equation}
where $\bm\epsilon\sim \mathcal{N}(\mathbf{0}, \mathbf{I})$, $\bm\epsilon_\theta$ is our conditional layout denoiser, $\bm{g}_t\sim\mathcal{N}(\bm{g}_t;\sqrt{\overline{\alpha}_t}\bm{g}_0, (1-\overline{\alpha}_t)\mathbf{I})$ is computed using the property of Gaussian distributions and $\bm{g}_0$ is the real layout geometric parameters. $\{\beta_t\}_{t=1}^T$ are forward process variances, $\alpha_t=1-\beta_t$ and $\overline{\alpha}_t=\prod_{s-1}^t\alpha_s$.

The training and sampling algorithm of LayoutDM are illustrated in \cref{alg:algorithm1} and \cref{alg:algorithm2} respectively.

\section{Experiments}
\label{sec:experiments}
\subsection{Experimental Settings}
\textbf{Datasets.} We evaluate our method on five public datasets of layouts for documents, natural scenes, magazines, text logos and mobile phone UIs.
\textbf{Rico}~\cite{Deka:2017:Rico} is a dataset of user interface designs for mobile applications, containing 72,219 user interfaces from 9,772 Android apps. \textbf{PublayNet}~\cite{zhong2019publaynet} contains 330K samples of machine-annotated scientific documents crawled from the Internet with five categories (text, title, figure, list, table). \textbf{Magazine}~\cite{10.1145/3306346.3322971} is a magazine layout dataset that covers a wide range of magazine categories. It consists of semantic layout annotations. \textbf{COCO}~\cite{10.1007/978-3-319-10602-1_48} is a large-scale labeled image dataset that contains images of natural scenes with instance segmentation. \textbf{TextLogo3K}~\cite{wang2021aesthetic} is a recently released dataset of text logos. It consists of 3,470 text logo images with manually annotated bounding boxes and pixel-level masks for each character.

\label{sec:evaluation metrics}
\textbf{Evaluation metrics.} 
To validate the effectiveness of LayoutDM, we employ four metrics in the literature to measure the perceptual quality of layouts. To be a fair comparison, we follow the guidance in~\cite{Kikuchi2021} to evaluate these metrics.
Specifically, \textbf{FID}~\cite{10.5555/3295222.3295408} measures the distribution distance between real layouts and generated layouts. We use the pre-trained classifiers in~\cite{Kikuchi2021} to compute FID. 
\textbf{Max. IoU}~\cite{Kikuchi2021} measures the similarity between the generated layouts and the reference layouts. It is designed to find the best match for each layout from the generated set to the reference set.
\textbf{Overlap} and \textbf{Alignment} measure the perceptual quality of generated layouts. Overlap measures the total overlapping area between any pair of bounding boxes inside the layout. Additionally, we measure the Alignment by computing an alignment loss proposed in~\cite{9106863}.

\textbf{Implementing details.}
The max time step $T$ in LayoutDM is set to 1000. We set the forward process variances to constants increasing linearly from $\beta_1=10^{-4}$ to $\beta_T=0.02$ following Ho's design ~\cite{NEURIPS2020_4c5bcfec}. We use eight transformer layers which use 8-head attention in our model. Adam optimizer~\cite{kingma2014adam} with a learning rate of $1\times 10^{-5}$ is used to optimize learnable parameters. The batch size is set to 1024. We implement our model with PyTorch~\cite{NEURIPS2019_9015} and PyTorch Lightning. All experiments are performed on a single NVIDIA Quadro RTX 6000 GPU device. 
\begin{table}[t]\footnotesize
\centering
\setlength\tabcolsep{1.6pt}
\begin{tabular}{ll|cccc}
\toprule[1.4pt]
               & Dataset         & \multicolumn{4}{c}{Rico}                                                           \\ \cline{3-6} 
Model          &                 & FID↓               & Max. IoU↑          & Alignment↓         & Overlap↓            \\ \hline
\multicolumn{2}{l|}{LayoutGAN-W~\cite{8948239}} & 162.75±0.28        & 0.30±0.00          & 0.71±0.00          & 174.11±0.22         \\
\multicolumn{2}{l|}{LayoutGAN-R~\cite{8948239}} & 52.01±0.62         & 0.24±0.00          & 1.13±0.04          & 69.37±0.66          \\
\multicolumn{2}{l|}{NDN-none~\cite{10.1007/978-3-030-58580-8_29}}    & 13.76±0.28         & 0.35±0.00          & 0.56±0.03          & \textbf{54.75±0.29} \\
\multicolumn{2}{l|}{LayoutGAN++~\cite{Kikuchi2021}} & 14.43±0.13         & 0.36±0.00          & 0.60±0.12          & 59.85±0.59          \\
\multicolumn{2}{l|}{VTN~\cite{9578374}} & 9.31±0.21         & 0.36±0.00          & 0.88±0.11          & 59.31±0.45          \\
\multicolumn{2}{l|}{LayoutDM(Ours)}        & \textbf{3.03±0.06} & \textbf{0.49±0.00} & \textbf{0.36±0.06} & 57.55±0.48          \\ \hline
\multicolumn{2}{l|}{Real data}   & 4.47               & 0.65               & 0.26               & 50.58               
\end{tabular}
\begin{tabular}{ll|cccc}
\toprule[1.4pt]
               & Dataset         & \multicolumn{4}{c}{PublayNet}                                                     \\ \cline{3-6} 
Model          &                 & FID↓               & Max. IoU↑          & Alignment↓         & Overlap↓           \\ \hline
\multicolumn{2}{l|}{LayoutGAN-W~\cite{8948239}} & 195.38±0.46        & 0.21±0.00          & 1.21±0.01          & 138.77±0.21        \\
\multicolumn{2}{l|}{LayoutGAN-R~\cite{8948239}} & 100.24±0.61        & 0.24±0.00          & 0.82±0.01          & 45.64±0.32         \\
\multicolumn{2}{l|}{NDN-none~\cite{10.1007/978-3-030-58580-8_29}}    & 35.67±0.35         & 0.31±0.00          & 0.35±0.01          & 16.5±0.29          \\
\multicolumn{2}{l|}{LayoutGAN++~\cite{Kikuchi2021}} & 20.48±0.29         & 0.36±0.00          & 0.19±0.00          & 22.80±0.32         \\
\multicolumn{2}{l|}{VTN~\cite{9578374}} & 13.07±0.47         & 0.37±0.00          & 0.30±0.01          & 13.15±0.24          \\
\multicolumn{2}{l|}{LayoutDM(Ours)}        & \textbf{4.04±0.08} & \textbf{0.44±0.00} & \textbf{0.15±0.00} & \textbf{3.73±0.08} \\ \hline
\multicolumn{2}{l|}{Real data}   & 9.54               & 0.53               & 0.04               & 0.22               
\end{tabular}
\begin{tabular}{ll|cccc}
\toprule[1.4pt]
               & Dataset         & \multicolumn{4}{c}{Magazine}                                                       \\ \cline{3-6} 
Model          &                 & FID↓               & Max. IoU↑          & Alignment↓         & Overlap↓            \\ \hline
\multicolumn{2}{l|}{LayoutGAN-W~\cite{8948239}} & 159.2±0.87         & 0.12±0.00          & \textbf{0.74±0.02}          & 188.77±0.93         \\
\multicolumn{2}{l|}{LayoutGAN-R~\cite{8948239}} & 100.66±0.35        & 0.16±0.00          & 1.90±0.02          & 111.85±1.44         \\
\multicolumn{2}{l|}{NDN-none~\cite{10.1007/978-3-030-58580-8_29}}    & 23.27±0.09         & 0.22±0.00          & 1.05±0.03          & \textbf{30.31±0.77} \\
\multicolumn{2}{l|}{LayoutGAN++~\cite{Kikuchi2021}} & 13.35±0.41         & 0.26±0.00          & 0.80±0.02 & 32.40±0.89          \\
\multicolumn{2}{l|}{VTN~\cite{9578374}} & 12.34±0.39         & 0.25±0.00          & 1.07±0.03          & 39.97±0.62          \\
\multicolumn{2}{l|}{LayoutDM(Ours)}        & \textbf{9.11±0.15} & \textbf{0.29±0.00} & 0.77±0.03          & 32.53±0.72          \\ \hline
\multicolumn{2}{l|}{Real data}   & 12.13              & 0.35               & 0.43               & 25.64               \\ \bottomrule[1.4pt]
\end{tabular}
\caption{Quantitative comparison conditioned on element category labels. 
For reference, the FID and Max. IoU computed between the validation and test data, and the Alignment and Overlap computed with the test data are shown as real data. 
LayoutGAN-W and LayoutGAN-R denote LayoutGAN with wireframe rendering discriminator and with relation-based discriminator.}
\label{tab:quantitative comparison}
\end{table}

\subsection{Quantitative Evaluation}
\label{sec:quantitative comparisons}

\textbf{Comparison with state-of-the-art models.} We quantitatively evaluate the quality of conditional layout generation results on benchmark datasets: Rico, PublayNet and Magazine. Since most methods are designed for unconditional layout generation and have no available public implements, we compare our method with conditional LayoutGAN~\cite{8948239}, NDN-none~\cite{10.1007/978-3-030-58580-8_29} and LayoutGAN++~\cite{Kikuchi2021} by citing the results in~\cite{Kikuchi2021}. Moreover, we also implement a conditional VTN~\cite{9578374} as our comparison baseline model, so that we can compare with methods based on both VAEs and GANs. 
All the metrics are computed on the full test splits of the datasets. We report the mean and standard deviation over five independent evaluations for each experiment.

The comparison results are reported in~\cref{tab:quantitative comparison}.  From this table, we can observe that: (1) Our method outperforms SOTA methods on both FID and Max. IoU on all three benchmark datasets, which indicates that the generated layouts by our method are more similar to the real layouts than those by SOTA methods. This verifies that our LayoutDM can produce higher-quality and more diverse layouts than SOTA methods, since FID captures both diversity and fidelity.
(2) Layouts generated by our method have a lower FID score than the real ones in validation splits. This is because the generated layouts have identical attributes to those in the test split, while the real layouts in the validation split have different attributes from those in the test split. We provide more analysis on this point in the supplementary material. (3) With regards to Alignment and Overlap, LayoutDM is slightly weaker on Rico and Magazine. Because our method lacks a discriminator that guides the generator to generate layouts with better alignment and overlap properties as in GANs and does not introduce the layout refine module in NDN. This is likely to cause LayoutDM to be inferior to the state-of-the-art on these two metrics.

Note here that, we don't compare LayoutDM to other state-of-the-art methods such as  LayoutTransformer~\cite{9710883} and Coarse2fine~\cite{Jiang2022CoarsetoFineGM}, because these methods focus on unconditional layout generation problem which is different to our setting. Recently, Kong \etal~\cite{Kong2021BLTBL} reimplement the conditional version of VTN and LayoutTransformer, and compare their proposed BLT model with these models under the settings of conditional layout generation. However, they compute the metrics (IoU~\cite{Kong2021BLTBL},  Alignment~\cite{10.1007/978-3-030-58580-8_29}, Overlap~\cite{8948239}) in different ways. As an extended comparison, we adopt the metrics used by BLT~\cite{Kong2021BLTBL}, and compare our LayoutDM with VTN, LayoutTransformer, LayoutVAE, NDN and BLT on PublayNet. The comparison results are reported in~\cref{tab:blt_metric}. As shown in this table, our model also outperforms SOTA methods on all metrics.

\begin{table}[t]\small
\setlength\tabcolsep{3.4pt}
\centering
\begin{tabular}{l|ccc}
\hline
                  & IoU↓~\cite{Kong2021BLTBL}           & Overlap↓~\cite{8948239}      & Alignment↓~\cite{10.1007/978-3-030-58580-8_29}    \\ \hline
L-VAE~\cite{9010959}             & 0.45\scriptsize±1.3\%     & 0.15\scriptsize±0.9\%    & 0.37\scriptsize±0.7\%    \\
NDN~\cite{10.1007/978-3-030-58580-8_29}               & 0.34\scriptsize±1.8\%     & 0.12\scriptsize±0.8\%    & 0.39\scriptsize±0.4\%    \\
VTN~\cite{9578374}               & 0.21\scriptsize±0.6\%     & 0.06\scriptsize±0.2\%    & 0.33\scriptsize±0.4\%    \\
Trans.~\cite{9710883} & 0.19\scriptsize±0.3\%     & 0.06\scriptsize±0.3\%    & 0.33\scriptsize±0.3\%    \\
BLT~\cite{Kong2021BLTBL}               & 0.19\scriptsize±0.2\%     & 0.04\scriptsize±0.1\%    & 0.25\scriptsize±0.7\%    \\
\textbf{Ours}          & \textbf{0.0053}\scriptsize±0.5\% & \textbf{0.01}\scriptsize±0.1\% & \textbf{0.22}\scriptsize±1.2\% \\ \hline
\end{tabular}
\caption{Comparison with extended methods on PublayNet. Qualitative results are cited from~\cite{Kong2021BLTBL}. ``Trans." denotes ``LayoutTransformer" and ``L-VAE" denotes ``LayoutVAE".}
\label{tab:blt_metric}
\end{table}

\begin{figure*}[t]
  \centering
   \includegraphics[width=\textwidth]{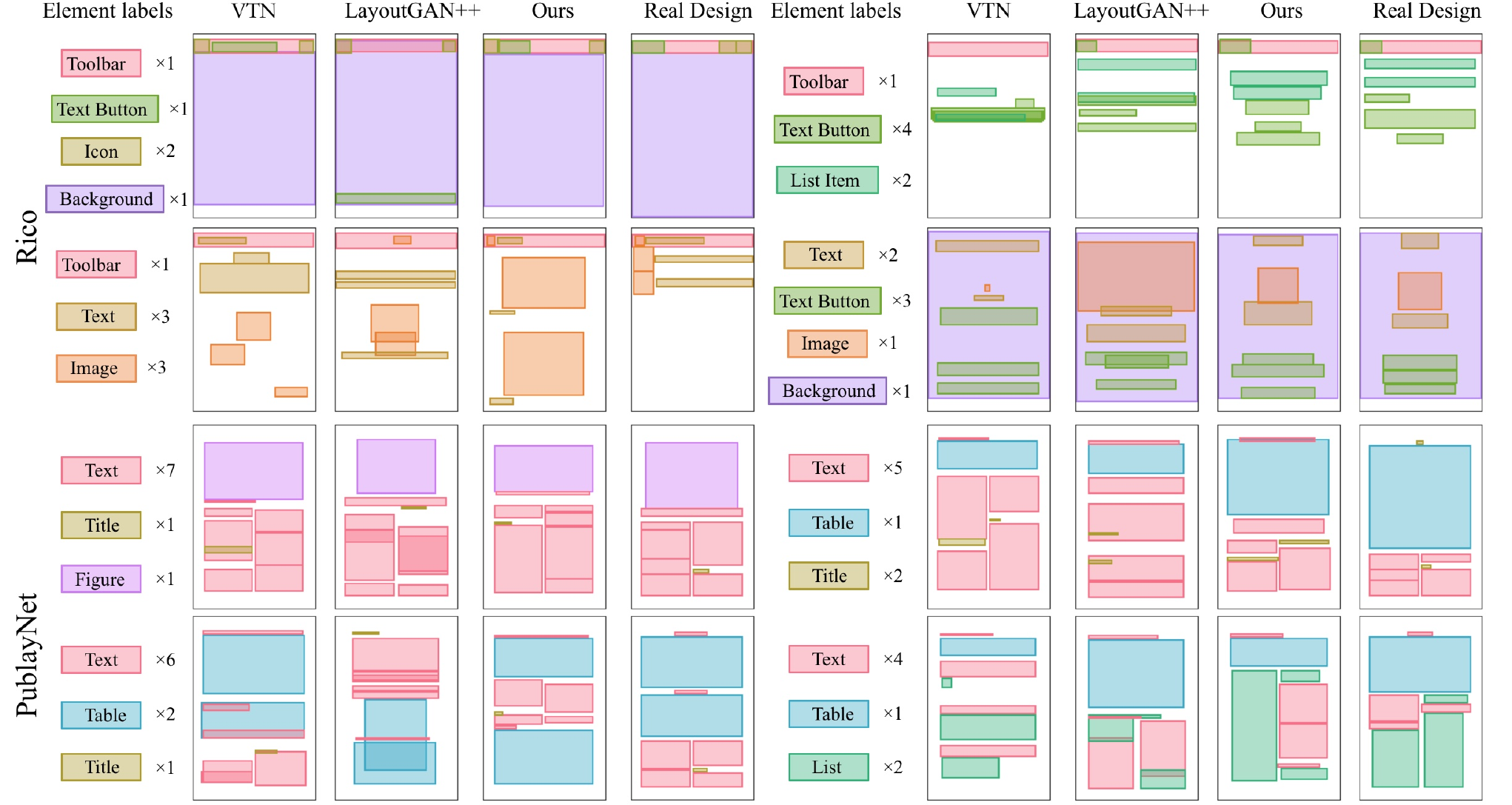}

   \caption{Qualitative comparison on Rico and PublayNet. Element labels indicate the labels of elements used as conditional inputs. 
   }
   \label{fig:diffusion_qualitative}
   \vspace{-0.2cm}
\end{figure*}

\begin{figure}
    \centering
    \includegraphics{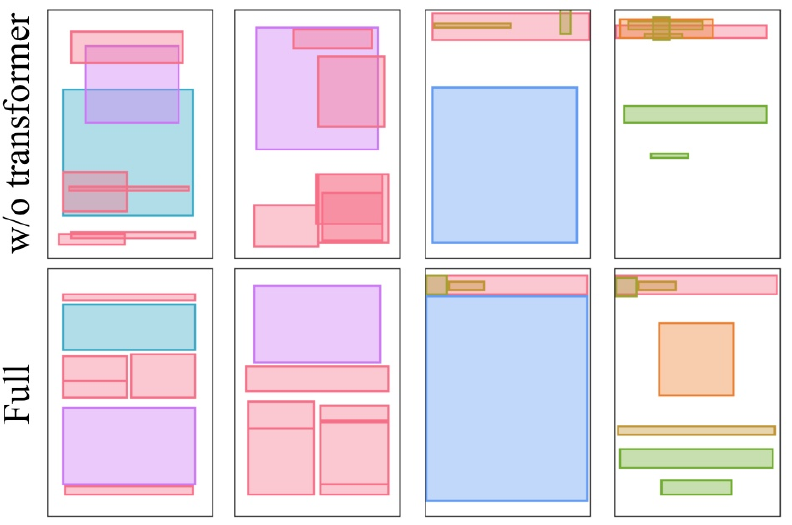}
    \caption{Ablation study on the effect of transformer layers. ``w/o transformer" denotes model without transformer layers. ``Full" denotes our full model.}
    \label{fig:ablation}
\end{figure}

\textbf{Effect of transformer layers.}~
We conduct ablation experiments to demonstrate the effectiveness of the transformer layers in LayoutDM. Quantitative results are reported in \cref{tab:ablation_transformer} and qualitative comparisons are shown in \cref{fig:ablation}. We have the following observation: After replacing the transformer structure in LayoutDM with a sequence of FC layers, the model can still predict a suitable size for each element, but the positional relationships between elements can not be handled, resulting in significant overlapping and misalignment. This proves that the transformer layers play an essential role and can efficiently capture and utilize the high-level relationships between elements to generate high-quality layouts which follow design rules and aesthetic principles.

\begin{table}[h]\small
\centering
\setlength\tabcolsep{4pt}
\begin{tabular}{lcccc}
\toprule
                & \multicolumn{4}{c}{Rico}                 \\ \cline{2-5} 
Architecture    & FID↓  & Max.IoU↑ & Alignment↓ & Overlap↓ \\ \hline
w/o transformer & 52.64 & 0.29     & 1.08       & 58.13    \\
Full model      & 3.03  & 0.49     & 0.36       & 57.55    \\ \toprule
                & \multicolumn{4}{c}{PublayNet}            \\ \cline{2-5} 
Architecture    & FID↓  & Max.IoU↑ & Alignment↓ & Overlap↓ \\ \hline
w/o transfomer  & 99.60 & 0.27     & 0.89       & 63.87    \\
Full model      & 4.04  & 0.44     & 0.15       & 3.73     \\ \bottomrule
\end{tabular}
\caption{The quantitative results of ablations on transformer layers. ``Full" denotes our full model. ``w/o transformer" denotes model without transformer layers.}
\label{tab:ablation_transformer}
\end{table}

\subsection{Qualitative Comparisons}
\textbf{Generation quality comparison.}
To qualitatively compare the generation performance of different models, we compare with the state-of-the-art method LayoutGAN++ and our implemented conditional VTN.    
We randomly sample layouts from the test dataset and use the element category labels as conditional inputs. \cref{fig:diffusion_qualitative} shows the qualitative comparison results. As one can see, LayoutDM can arrange elements in a reasonable and complicated way, generating higher quality layouts than the other two, with fewer overlapping and better alignment. 

\begin{figure}[t]
    \centering
    \includegraphics[width=\linewidth]{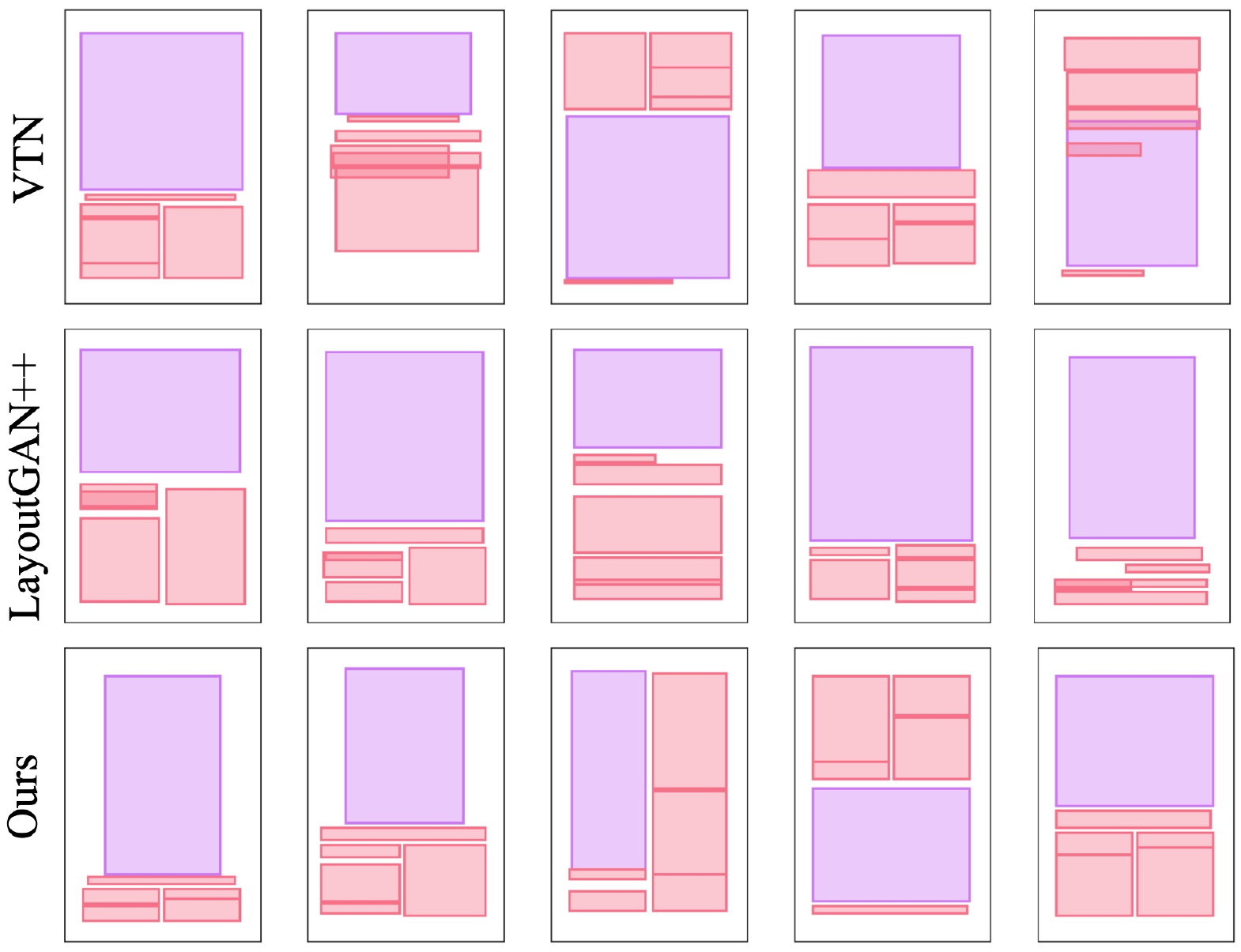}
    \caption{Diversity comparison on PublayNet. We show five samples generated by giving the same category attributes as condition: one \textit{Figure} and five \textit{Texts}. The \textit{Figure} is drawn in purple and the \textit{Texts} are drawn in red.}
    \label{fig:Diversity Compare}
\end{figure}

\textbf{Generation diversity comparison.}~The diversity of results is also an important factor in evaluating the method. We compare the diversity of layouts generated by conditional VTN, LayoutGAN++, and LayoutDM. \cref{fig:Diversity Compare} show the comparison results. One can see that conditional VTN and LayoutDM generate more diverse results. The \textit{Figure} element in the results floats on the page. Compared to the other two models, LayoutGAN++ captures less diversity, which places a large \textit{Figure} element on the top of the pages in columns 2,4, and 5. LayoutDM performs well in diversity because it breaks the generation into a series of conditional diffusion steps which are relatively easy to model. This alleviates the mode collapse problem in strongly conditional generation task that can lead to the generation of similar modes. 
We provide more qualitative comparisons on diversity in the supplementary material.

\begin{figure}[t]
    \centering
    \includegraphics[width=\linewidth]{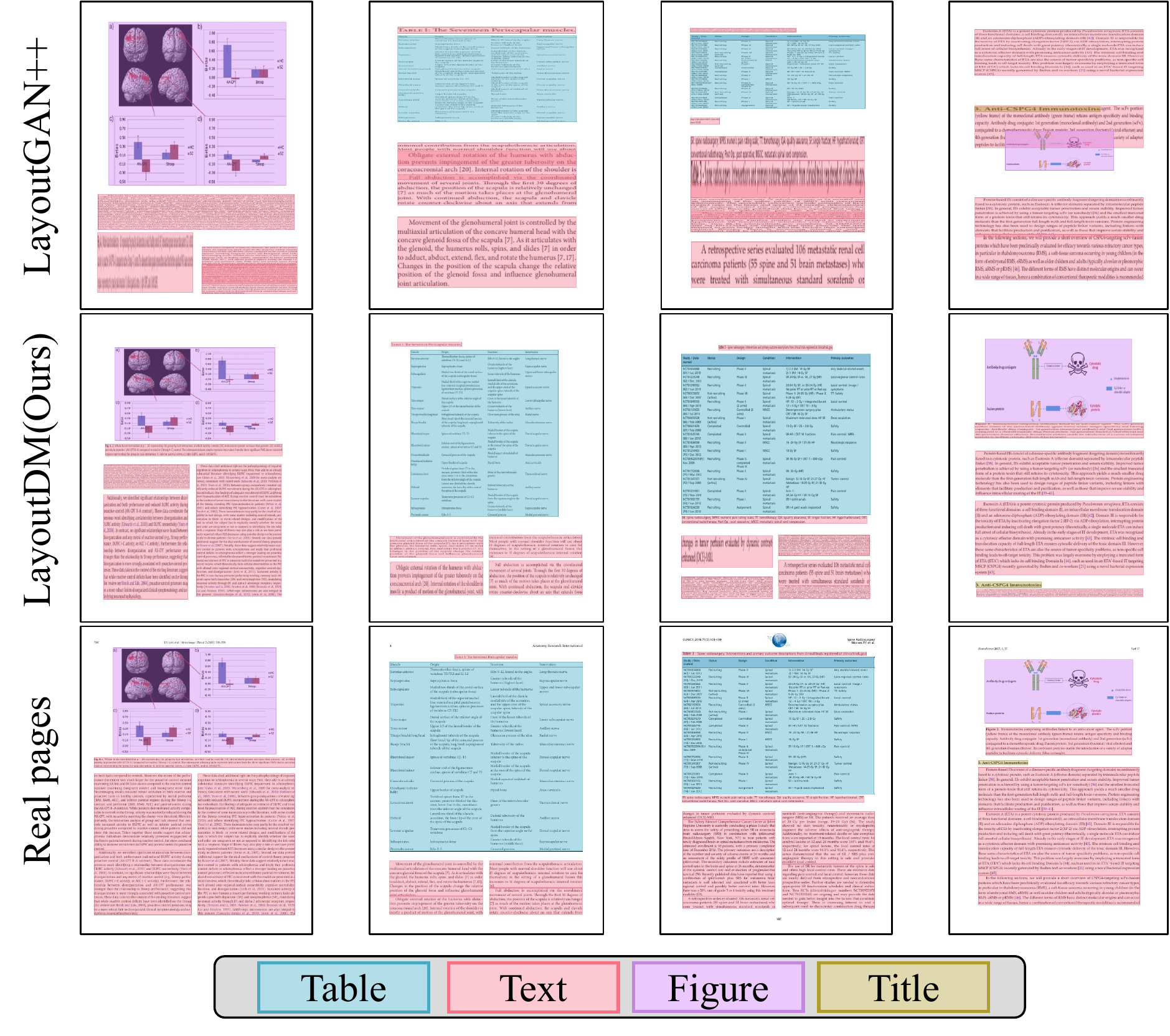}
    \caption{Rendering results comparison. Top: Rendered pages using layouts generated by LayoutGAN++. Middle: Rendered pages using layouts generated by LayoutDM. Bottom: Real paper pages in PublayNet.}
    \label{fig:diffusion render}
\end{figure}

\textbf{Rendering results comparison.}~We render graphic pages for better visualization using the generated layouts. \cref{fig:diffusion render} show the rendering results comparison on PublayNet.
We find the layouts generated by LayoutDM follow design rules well and reasonably allocate page space. Compared to the results generated by LayoutGAN++, our results are better in alignment and have no overlapping between elements. Note that we crop the elements from the original document and then render the pages using a simple resize-to-fit method, so the text areas and figures will suffer some distortion. In real design scenarios, this problem can be solved by element customization (\eg, adjust the font size).

\subsection{Extended Layout Generation Tasks}

\begin{figure}[t]
    \centering
    \includegraphics[width=\linewidth]{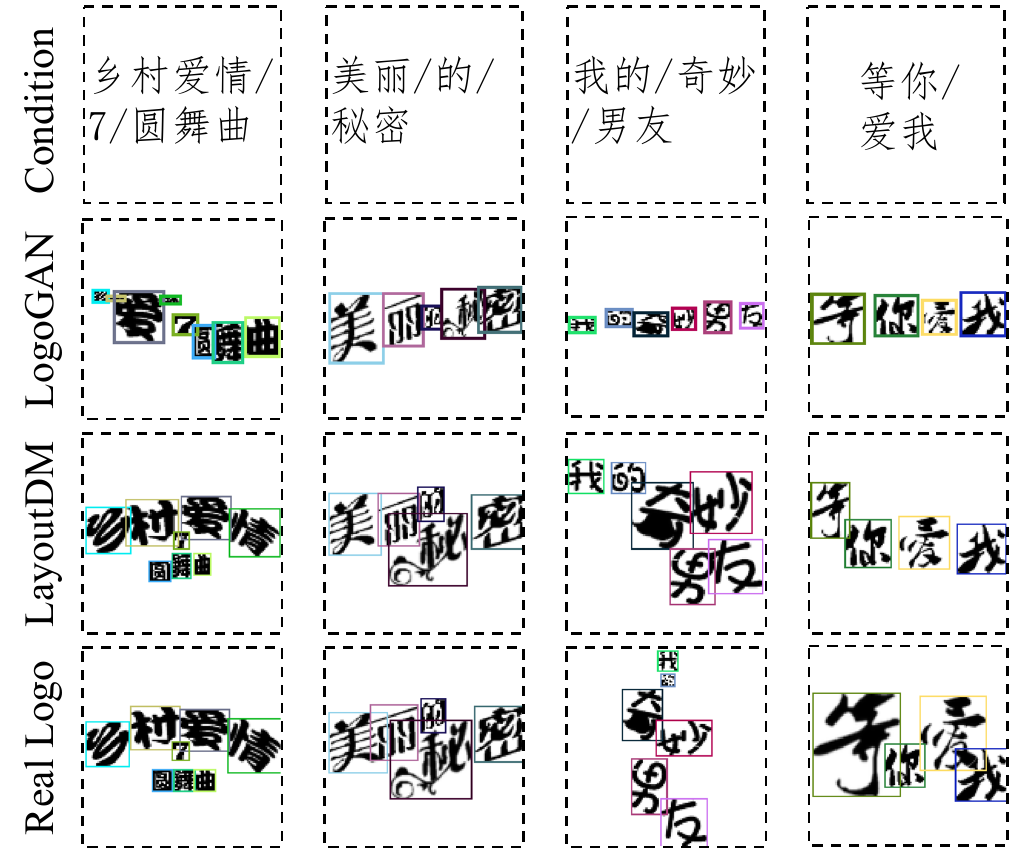}

    \caption{Text logo generation results. ``LogoGAN" denotes the text logo generation model proposed in~\cite{wang2021aesthetic}. ``Condition" represents the textual features (including character and word embeddings) used as conditional input in LogoGAN and LayoutDM. ``/" is the symbol for splitting tokens.}
    \label{fig:diffusion logo}
\end{figure}

\textbf{Text logo layout generation.}~
TextLogo3K~\cite{wang2021aesthetic} dataset contains character and word embedding of texts in the logos. Although the dataset does not provide any label information, we can still generate  logo layouts  conditioned on the provided textual features. Positional encoding is added to LayoutDM to make the transformer structure aware of the reading order in textual feature sequences. We compare the logo layout generation results of LayoutDM and those generated by the logo layout generator provided by TextLogo3K~\cite{wang2021aesthetic}. The qualitative results are shown in \cref{fig:diffusion logo}. As one can see, our model generates reasonable logo layouts while maintaining the correct reading order and aesthetic principles. Compared to LogoGAN, the style of the logos generated by our model is more flexible, not simply arranging the text from left to right. Our model also performs better when there are large numbers of characters in the layout, where LogoGAN fails to generate reasonable results.


\begin{figure}[t]
    \centering
    \includegraphics[width=\linewidth]{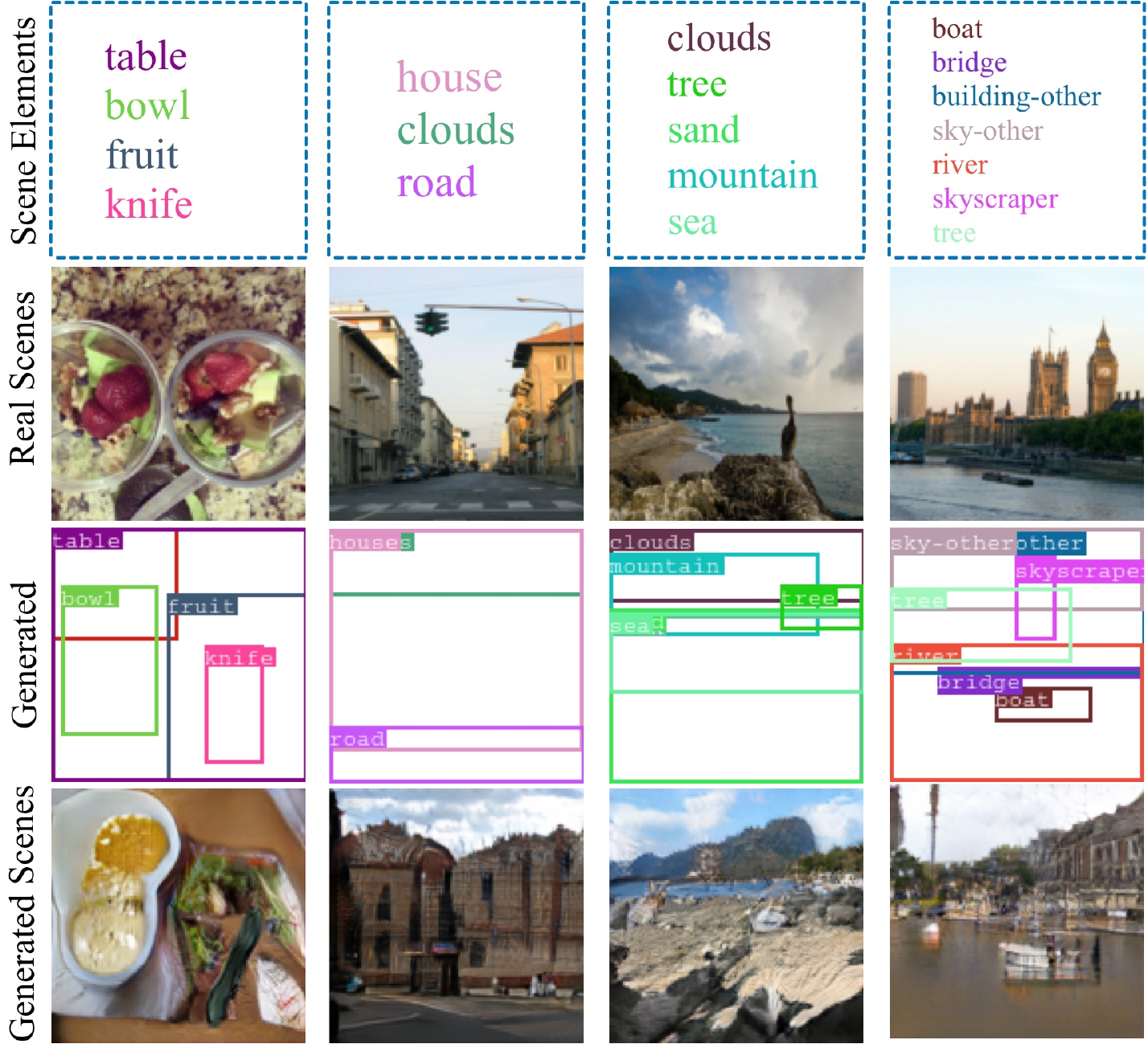}

    \caption{Natural scene layout generation results on COCO. Our model uses scene element labels as conditional input to generate reasonable scene layouts. 
    }
    \label{fig:diffusion coco}
\end{figure}

\textbf{Scene layout generation.}~Our model can also generate natural scene layouts. We illustrate the qualitative results of the scene layout generation on COCO in \cref{fig:diffusion coco}. Given the labels of scene elements in a scene, our model generates reasonable scene layout proposals. We then use
a downstream layout-to-image generation application~\cite{9010748} to finally render natural scene images. The results show that our model learns the principle of scene layouts well and can understand the complex relationships between elements in natural scenes. For example, the boat should be in the river and the cloud should be in the upper part of the scene.

\subsection{Limitations}
Although our method shows impressive results in the conditional layout generation problem in comparison to existing methods, it still has limitations. For example, like other layout generation methods, our approach treats design elements as being on a single-layer canvas. This can not model a layout with multiple layers occluding each other. Our method also has no advantage over other generative models in generation speed because the generation of the diffusion model requires an iterative denoising process. We leave the solution to the above problems for future work.

\section{Conclusion}
\label{sec:discussion}
This paper proposes a transformer-based diffusion model LayoutDM to address conditional layout generation. We introduce a purely transformer-based Layout Denoiser to model the diffusion reverse process. Benefitting from both DDPM and transformer, in comparison to existing methods, LayoutDM can generate high-quality generation with desired properties such as better diversity, faithful distribution coverage, and stationary training. Quantitative and qualitative results demonstrate that our model outperforms the state-of-the-art methods in terms of visual perceptual quality and diversity.

\noindent 
\textbf{Acknowledgment.} This work was supported in part to Dr. Liansheng Zhuang by NSFC under contract No.U20B2070 and No.61976199, in part to Dr. Fengying Yan by NSFC under contract No.42341207.

{\small
\bibliographystyle{ieee_fullname}
\bibliography{egbib}
}

\end{document}


\title{Supplementary Material for LayoutDM}
\date{}
\maketitle

Due to limited space, our paper does not show all the comparison results. In this document, we will give more details about our experiments and present additional results.
\section{Datasets}
We summarize the statistics of the datasets used in our paper in Table \ref{tb:dataset_statistics}. In all our experiments, we follow the following settings for training, validation and testing. For PublayNet, COCO, and TextLogo3K, we use 95\% of the official training split for training, the rest for validation, and the official validation split for testing. For Rico and Magazine, we use 85\% of the dataset for training, 5\% for validation, and 10\% for testing. 

\begin{table}[htb]
\centering
\setlength\tabcolsep{3.8pt}
\begin{small}
\begin{tabular}{lccrrr}
\hline
Dataset      &\begin{tabular}[c]{@{}c@{}}\# label \\ types\end{tabular} & \begin{tabular}[c]{@{}c@{}}Max. \\ \# elements\end{tabular} & \# train.   & \# val.   & \# test.  \\ \hline
Rico      & 13     & 9                                                       & 17,515  & 1,030 & 2061  \\
PublayNet & 5      & 9                                                       & 160,549 & 8,450 & 4,226 \\
Magazine  & 5      & 33                                                      & 3,331   & 196   & 392   \\
COCO      & 183    & 8                                                       & 71,038  & 3,739 & 3,097 \\
TextLogo3K      & N/A    & 20                                                      & 3,011   & 159   & 300   \\ \hline
\end{tabular}
\end{small}
\caption{Statistics of the datasets used in our experiments.}
\label{tb:dataset_statistics}
\end{table}

\section{About Evaluation Metrics}
Previous work often uses different metrics to evaluate the visual quality and realism of the generated layouts. Though some metrics have the same name (such as Alignment and Overlap), they are computed in different ways. In our paper, we use the metrics in~\cite{Kikuchi2021} to evaluate our method, which is different from those in BLT~\cite{Kong2021BLTBL}. Here, we detail how to compute these metrics. 

\subsection{Metrics in our paper}
In our paper, we follow the guidance in~\cite{Kikuchi2021} to evaluate our proposed method. Four evaluation metrics are used to evaluate the quality of the generated layouts: FID~\cite{Kikuchi2021}, Max. IoU, Alignment~\cite{9106863} and Overlap~\cite{9106863}. 

\textbf{FID}~\cite{Kikuchi2021} measures the realism and accuracy of the generated layouts. To compute FID, we train a binary layout classifier to discriminate between real layouts and noise added layouts, and use the intermediate features of the network as the representative features of layouts.  For a fair comparison, we use the pre-trained neural network in~~\cite{Kikuchi2021} to calculate the FID metrics.

\textbf{Max. IoU}~\cite{Kikuchi2021} is defined between two collections of generated layouts and references. We compute the Max. IoU metric as in~\cite{Kikuchi2021}. In our paper, the layout geometric sequence $\bm{g}=(g_1, \cdots, g_N)$  corresponds to the layout $B=\{b_i\}^{N}_{i=1}$ in~\cite{Kikuchi2021}, and the layout attributes sequence $\bm{f}=(f_1, \cdots, f_N)$ corresponds to the label set $l_i, i=1,\cdots, N$ in ~\cite{Kikuchi2021}.

\textbf{Alignment}~\cite{9106863} computes an alignment loss with the intuition that objects in graphic design are often aligned either by center or edge. Denote $\bm\theta=(x^L, y^T, x^C, y^C, x^R, y^B)$ as the top-left, center and bottom-right coordinates of the bounding box, we calculate the Alignment loss for each layout as follows: 

$$
L_{a l g}=\frac{1}{N}\sum_{i=1}^{N} \min \left(\begin{array}{l}
\boldsymbol{g}\left(\Delta x_{i}^{L}\right), \boldsymbol{g}\left(\Delta x_{i}^{C}\right), \boldsymbol{g}\left(\Delta x_{i}^{R}\right) \\
\boldsymbol{g}\left(\Delta y_{i}^{T}\right), \boldsymbol{g}\left(\Delta y_{i}^{C}\right), \boldsymbol{g}\left(\Delta y_{i}^{B}\right)
\end{array}\right)
$$
where $\bm{g}(x)=-\log(1-x)$, $N$ is the number of elements in the layout, and $\Delta x_{i}^{*}(*=L, C, R)$ is computed as:
$$
\Delta x_{i}^{*}=\min _{\forall j \neq i}\left|x_{i}^{*}-x_{j}^{*}\right|
$$
$\Delta y_{i}^{*}(*=T, C, B)$ can be computed similarly. The final value is multiplied by 100$\times$ for visibility following~\cite{Kikuchi2021}.

\textbf{Overlap}~\cite{9106863} measures the total overlapping area between any pair of bounding boxes in a layout. We compute the Overlap metric for each layout as follows:
$$
L_{o v e r}=\frac{1}{N}\sum_{i=1}^{N} \sum_{\forall j \neq i} \frac{s_{i} \cap s_{j}}{s_{i}}
$$where $s_{i} \cap s_{j}$ denotes the overlapping area between element $i$ and $j$. $N$ is the number of elements in the layout. The final value is multiplied by 100$\times$ for visibility following~\cite{Kikuchi2021}.

\subsection{Metrics in BLT}

In this paper, we also follow the guidance of BLT~\cite{Kong2021BLTBL} to evaluate our proposed method. The evaluation metrics used in BLT~\cite{Kong2021BLTBL} are different from those in~\cite{Kikuchi2021}: IoU, Alignment, and Overlap.

\begin{figure}[h]
    \centering
    \includegraphics[width=\linewidth]{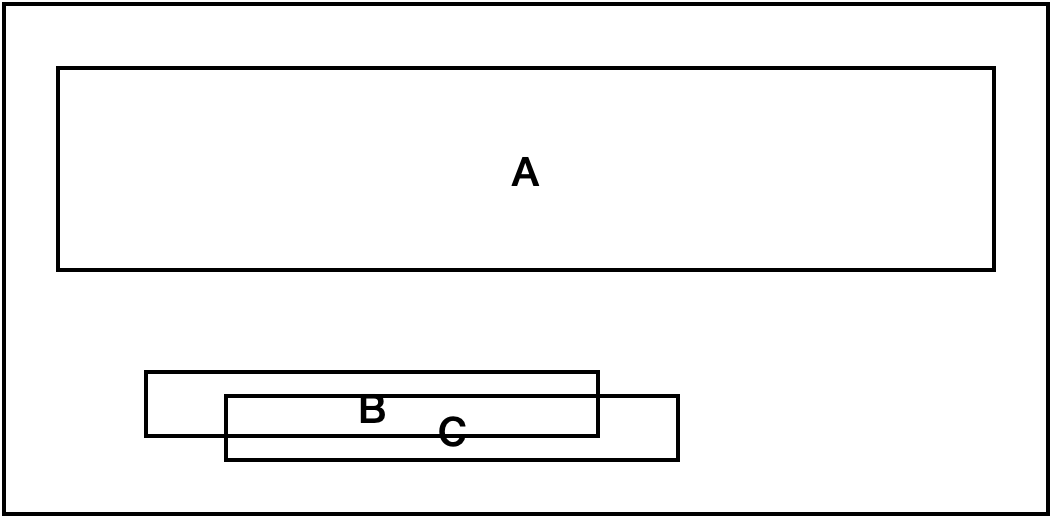}
    \caption{A toy layout sample for the IoU computation. The areas of objects A, B and C are 5, 1, 1, the overlapped area of B and C are 0.5. IoU $\frac{0.5}{6.5}=\frac{1}{13}$.}
    \label{fig:toy_sample}
\end{figure}

\textbf{IoU}~\cite{Kong2021BLTBL} measures the intersection over the union beween the generated bounding boxes. Following~\cite{Kong2021BLTBL}, an improved perceptual IOU is used. It first projects the layouts as if they were images then computes the ovarlapped area divided by \textbf{the union area of all objects}. A toy sample images is presented in \cref{fig:toy_sample} \footnote{Metric description and sample image are cited from the appendix of~\cite{Kong2021BLTBL}}

\textbf{Alignment}~\cite{10.1007/978-3-030-58580-8_29} measures the alignment among design elements. The average Alignment on a collection of layouts is computed using:
$$\frac{1}{N_{D}} \sum_{d} \sum_{i} \min _{j, i \neq j}\left\{\min \left(l\left(c_{i}^{d}, c_{j}^{d}\right), m\left(c_{i}^{d}, c_{j}^{d}\right), r\left(c_{i}^{d}, c_{j}^{d}\right)\right\}\right)$$where $N_D$ is the number of generated layouts, $c_k^d$ is the $k$-th component of the $d$-th layout. In addition, $l$, $m$, and $r$ are alignment functions where the distances between the left, center, and right of elements are measured, respectively. ~\cite{Kong2021BLTBL} uses L1 distance as the distance measure but does not adopt the function $\bm{g}(x)=-\log(1-x)$ to process the distance. ~\cite{Kong2021BLTBL} also 
do not normalize the Alignment by the number of elements $N$ or multiply the value by $100\times$. This leads to different results using the two different Alignments.

\textbf{Overlap}~\cite{8948239} is the percentage of total overlapping area among any two bounding boxes inside the whole page. 
$$
L_{o v e r}=\sum_{i=1}^{N} \sum_{\forall j \neq i} \frac{s_{i} \cap s_{j}}{s_{i}}
$$
~\cite{Kong2021BLTBL} does not normalize the values with the number of elements $N$ or multiply the value by $100\times$. This leads to different results using the two different Overlaps.


\section{Comparison with BLT}
BLT~\cite{Kong2021BLTBL} is another method for conditional layout generation, which is recently published in ECCV2022. Due to limited space, we only compare our LayoutDM with BLT on the PublayNet dataset. Here, we show more comparison results on more datasets using the metrics of BLT under the setting of conditional layout generation. The SOTA methods being compared include LayoutVAE~\cite{9010959}, LayoutTransformer~\cite{9710883}, VTN~\cite{9578374}, NDN~\cite{10.1007/978-3-030-58580-8_29}, LayoutGAN++~\cite{Kikuchi2021} and BLT~\cite{Kong2021BLTBL}. Since most SOTA methods (except LayoutGAN++) have no available public implements, we directly cite the experimental results in~\cite{Kong2021BLTBL}. All the experiments run five times, and the mean and standard deviation over five trials are reported. 
The full comparison results are shown in~\cref{tab:BLT}. As we can see, our LayoutDM significantly outperforms SOTA methods on PublayNet, Rico, and Maganize. 

\begin{table*}[htbp]
\centering
\begin{tabular}{llccccccc}
\hline
           & Dataset                   & \multicolumn{3}{c}{PublayNet}                                              & \multicolumn{3}{c}{Rico}                                                 & Magazine         \\ \hline
Model      & \multicolumn{1}{l|}{}     & IoU↓           & Overlap↓       & \multicolumn{1}{c|}{Alignment↓}          & IoU↓          & Overlap↓      & \multicolumn{1}{c|}{Alignment↓}          & IoU↓             \\ \hline
\multicolumn{2}{l|}{L-VAE~\cite{9010959}}             & 0.45\scriptsize±1.3\%     & 0.15\scriptsize±0.9\%     & \multicolumn{1}{c|}{0.37\scriptsize±0.7\%}          & 0.41\scriptsize±1.5\%    & 0.39\scriptsize±2.3\%    & \multicolumn{1}{c|}{0.38\scriptsize±1.9\%}          & \textbackslash{} \\
\multicolumn{2}{l|}{NDN~\cite{10.1007/978-3-030-58580-8_29}}               & 0.34\scriptsize±1.8\%     & 0.12\scriptsize±0.8\%     & \multicolumn{1}{c|}{0.39\scriptsize±0.4\%}          & 0.37\scriptsize±1.7\%    & 0.36\scriptsize±1.9\%    & \multicolumn{1}{c|}{0.41\scriptsize±1.6\%}          & \textbackslash{} \\
\multicolumn{2}{l|}{VTN~\cite{9578374}}               & 0.21\scriptsize±0.6\%     & 0.06\scriptsize±0.2\%     & \multicolumn{1}{c|}{0.33\scriptsize±0.4\%}          & 0.30\scriptsize±0.1\%    & 0.30\scriptsize±0.3\%    & \multicolumn{1}{c|}{0.32\scriptsize±0.9\%}          & 0.18\scriptsize±1.8\%       \\
\multicolumn{2}{l|}{Trans.~\cite{9710883}} & 0.19\scriptsize±0.3\%     & 0.06\scriptsize±0.3\%     & \multicolumn{1}{c|}{0.33\scriptsize±0.3\%}          & 0.31\scriptsize±0.2\%    & 0.33\scriptsize±0.8\%    & \multicolumn{1}{c|}{0.30\scriptsize±0.8\%}          & 0.20\scriptsize±0.8\%       \\
\multicolumn{2}{l|}{BLT~\cite{Kong2021BLTBL}}               & 0.19\scriptsize±0.2\%     & 0.04\scriptsize±0.1\%     & \multicolumn{1}{c|}{0.25\scriptsize±0.7\%} & 0.30\scriptsize±0.4\%    & 0.23\scriptsize±0.2\%    & \multicolumn{1}{c|}{\textbf{0.20}\scriptsize±1.1\%} & 0.18\scriptsize±0.6\%       \\
\multicolumn{2}{l|}{LayoutGAN++~\cite{Kikuchi2021}}               & 0.084\scriptsize±1.1\%     & 0.05\scriptsize±0.3\%     & \multicolumn{1}{c|}{0.24\scriptsize±0.5\%}          & \textbf{0.22}\scriptsize±1.2\%    & 0.18\scriptsize±0.85\%    & \multicolumn{1}{c|}{0.25\scriptsize±0.6\%}          & 0.061\scriptsize±1.7\%       \\
\multicolumn{2}{l|}{LayoutDM(Ours)}          & \textbf{0.0053}\scriptsize±0.5\% & \textbf{0.01}\scriptsize±0.1\% & \multicolumn{1}{c|}{\textbf{0.22}\scriptsize±1.2\%}                & \textbf{0.22}\scriptsize±0.7\% & \textbf{0.17}\scriptsize±0.7\% & \multicolumn{1}{c|}{0.24\scriptsize±1.2\%}                & \textbf{0.055}\scriptsize±0.9\%   \\ \hline
\multicolumn{2}{l}{Real Data}          & 0.00097        & 0.0006         & 0.283                                    & 0.19          & 0.167         & 0.277                                    & 0.054            \\ \hline
\end{tabular}
\caption{Quantitative comparison using the evaluation metrics in BLT. ``Trans." and ``L-VAE" denote ``LayoutTransformer" and ``LayoutVAE". The metrics computed with the test datasets are shown as Real Data.}
\label{tab:BLT}
\end{table*}

\begin{table*}[]
    \centering
    \begin{tabular}{ll|cccc}
    \toprule[1.4pt]
                      & Dataset            & \multicolumn{4}{c}{Rico}                         \\ \hline
    Model             &                    & FID↓        & Max. IoU↑  & Alignment↓  & Overlap↓    \\ \hline
    \multicolumn{2}{l|}{LayoutDM/PE (shuffled)}   & 14.97±0.48 & 0.35±0.00 & 0.42±0.05 & 55.65±0.31 \\
    \multicolumn{2}{l|}{LayoutDM/PE (original)}   & 3.26±0.08 & 0.48±0.01 & 0.32±0.05 & 57.45±0.27 \\
    \multicolumn{2}{l|}{LayoutDM (shuffled)} & 4.04±0.10  & 0.46±0.00 & 0.29±0.02  & 58.32±0.23 \\ 
    \multicolumn{2}{l|}{LayoutDM (original)} & 3.95±0.09  & 0.46±0.00 & 0.28±0.04  & 58.59±0.40 \\ \bottomrule[1.4pt]
    \end{tabular}
    
    \begin{tabular}{ll|cccc}
                      & Dataset            & \multicolumn{4}{c}{PublayNet}                      \\ \hline
    Model             &                    & FID↓        & Max. IoU↑   & Alignment↓   & Overlap↓    \\ \hline
    \multicolumn{2}{l|}{LayoutDM/PE (shuffled)}   & 29.83±0.54 & 0.34±0.01 & 0.28±0.01 & 16.03±0.27 \\
    \multicolumn{2}{l|}{LayoutDM/PE (original)}   & 3.96±0.08 & 0.44±0.00 & 0.14±0.00 & 3.93±0.09 \\
    \multicolumn{2}{l|}{LayoutDM (shuffled)} & 4.15±0.10  & 0.44±0.00 & 0.14±0.01 & 4.23±0.07 \\
    \multicolumn{2}{l|}{LayoutDM (original)} & 4.17±0.12  & 0.44±0.00  & 0.14±0.01  & 4.13±0.05  \\ \bottomrule[1.4pt]
    \end{tabular}
    
    \begin{tabular}{ll|cccc}
                      & Dataset            & \multicolumn{4}{c}{Magazine}                             \\ \hline
    Model             &                    & FID↓                 & Max. IoU↑ & Alignment↓ & Overlap↓    \\ \hline
    \multicolumn{2}{l|}{LayoutDM/PE (shuffled)}   & 19.23±0.54 & 0.24±0.00 & 0.89±0.01 & 54.59±1.35 \\
    \multicolumn{2}{l|}{LayoutDM/PE (original)}   & 9.44±0.13 & 0.29±0.00 & 0.84±0.02 & 35.64±0.68 \\
    \multicolumn{2}{l|}{LayoutDM (shuffled)} & 10.05±0.19          & 0.28±0.00 & 0.78±0.04 & 33.76±0.81 \\
    \multicolumn{2}{l|}{LayoutDM (original)} & 9.75±0.40           & 0.29±0.00 & 0.79±0.02 & 32.74±0.58 \\ \bottomrule[1.4pt]
    \end{tabular}
    \caption{Quantitative results on the effect of element order.}
    \label{tab:order}
\end{table*}

\section{Why LayoutDM has lower FID value than ``Real Data"?}
The reason lies in the definition of FID metric. According to \cite{Kikuchi2021}, the FID value measures the distribution distance between the layouts generated by the models and the real layouts in \emph{test set}. For ``real data", the FID value is computed using the real layouts in \emph{validation set} and those in \emph{test set}. Since we generate the layouts conditioned on the attributes of layouts in test set, the distribution of generated layouts should be more similar to that of the test set than validation set. So, our method has a lower FID value than ``real data". 

Moreover, to verify our analysis, we generate layouts conditioned on the layout attributes in the validation set, and compute the FID value of our method (denoted as ``Gen./Val. Set" in the following table). The result is shown in \cref{tab:FID}. In this case, we can observe that the FID value of our method is close to that of ``Real data'', which is consistent with our analysis. 

\begin{table}[ht]
\centering
\begin{tabular}{l|c|c|c}
\hline
\hline
     & Gen. / Test set & \textbf{Gen. / Val. set} & Real data \\ \hline
FID↓ & 4.04±0.08            & \textbf{10.86±0.04}               & 9.54      \\ \hline
\end{tabular}
\caption{FID comparison on PublayNet. Gen. denotes ``Generated" and Val. denotes ``Validation".}
\label{tab:FID}
\end{table}

\section{Ablation study on positional encoding.}
\textbf{Qualitative results.} In this paper, we don't consider the order of designed elements on a canvas, so our LayoutDM omits the positional encoding in its transformer component. To evaluate the impact of positional encoding, we train a LayoutDM model with positional encoding as a reference (denoted as LayoutDM/PE). We randomly shuffle the order of elements in the input sequence in all three data sets, and evaluate the LayoutDM and LayoutDM/PE in the shuffled sequence.

\begin{figure*}[h]
    \centering
    \includegraphics[width=0.8\textwidth]{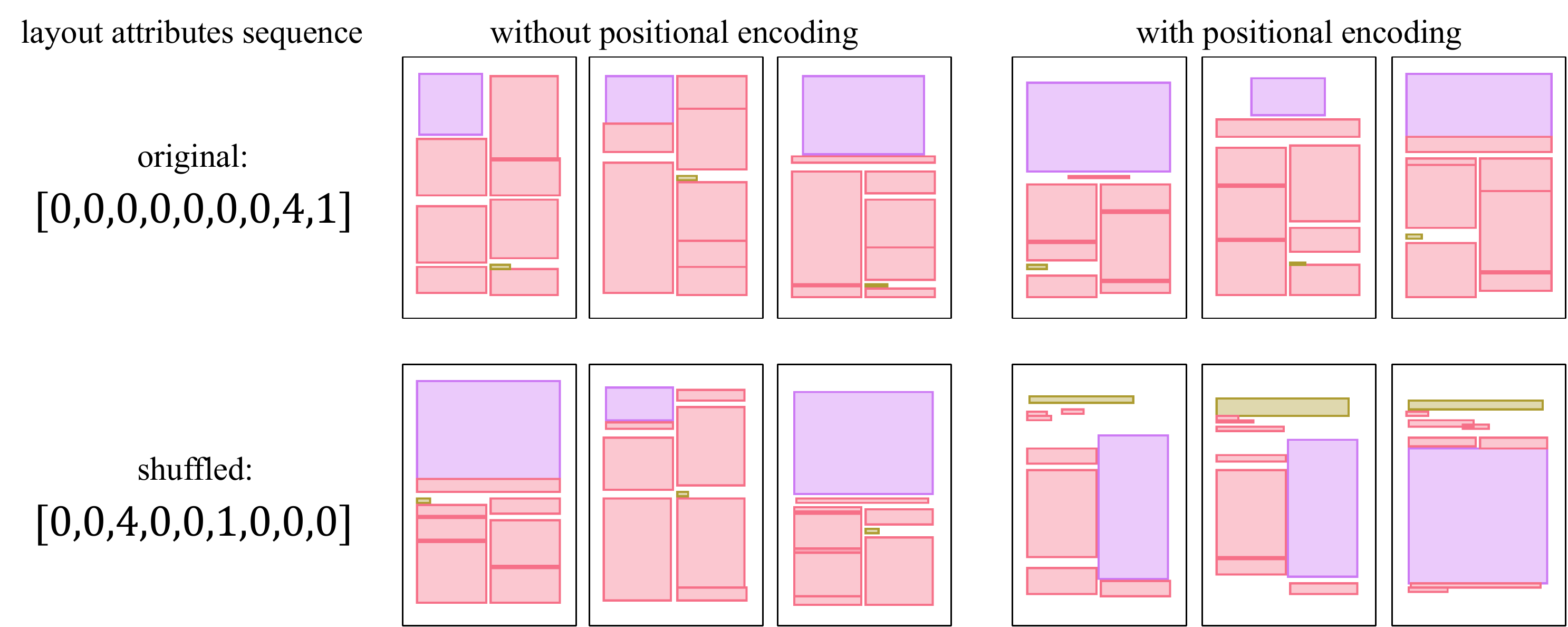}
    \caption{Effect of element order of input sequence. We use index ``0" , ``1" and ``4" to represent element category label ``Text", ``Title" and ``Figure".}
    \label{fig:order}
    \includegraphics[width=\textwidth]{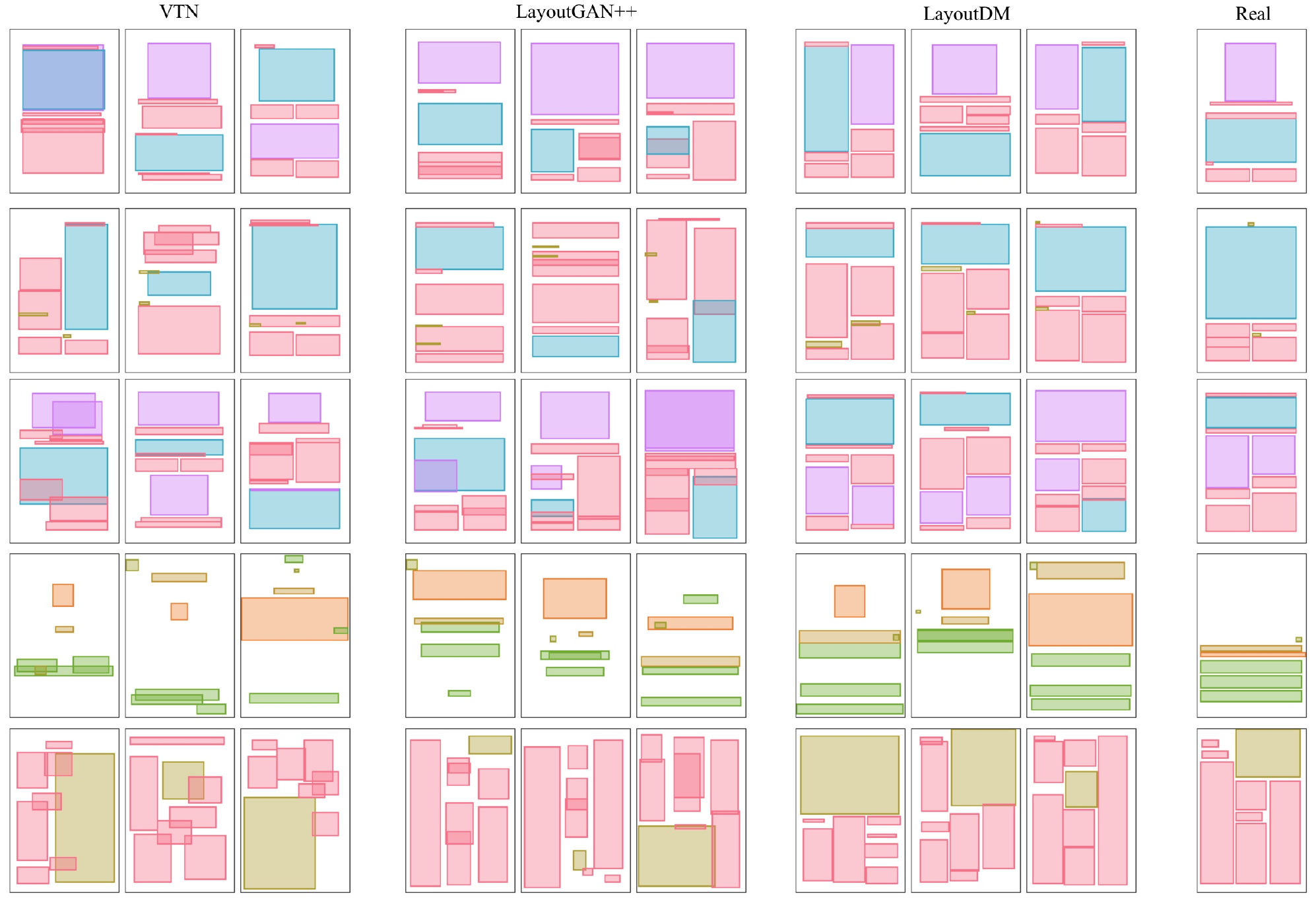}
    \caption{Additional conditional generation comparison. We show three samples for each model conditioned on the same element category labels. Real design are shown as Real.}
    \label{fig:conditional}
\end{figure*}

The results are shown in \cref{fig:order}. As we can see, after removing the positional encoding, the model is no longer aware of the positional information in the input sequence, and thus can still generate reasonable results conditioned on the shuffled input sequence. If we use positional encoding in our model, the model (LayoutDM/PE) will interpret the original sequence and the shuffled sequence as completely different inputs and thus produces results of bad quality.

\textbf{Quantitative results.} \Cref{tab:order} shows the quantitative results. As one can see, shuffling the order of elements does not affect the model without positional encoding, while the performance of the model with positional encoding is significantly affected. This proves that the performance of our model is independent of the order of the elements in the input sequence.

\section{More Qualitative Results}
This section presents more qualitative results. We show the comparison results of three models in \cref{fig:conditional}, namely our implemented conditional VTN, LayoutGAN++ and LayoutDM. As one can see, our model has higher quality in terms of alignment and overlap than the other two methods. The layouts generated by our LayoutDM are very similar to the real data. 

\section{More Qualitative Comparisons on Diversity}
\begin{figure*}[h]
    \centering
    \includegraphics[width=0.95\textwidth]{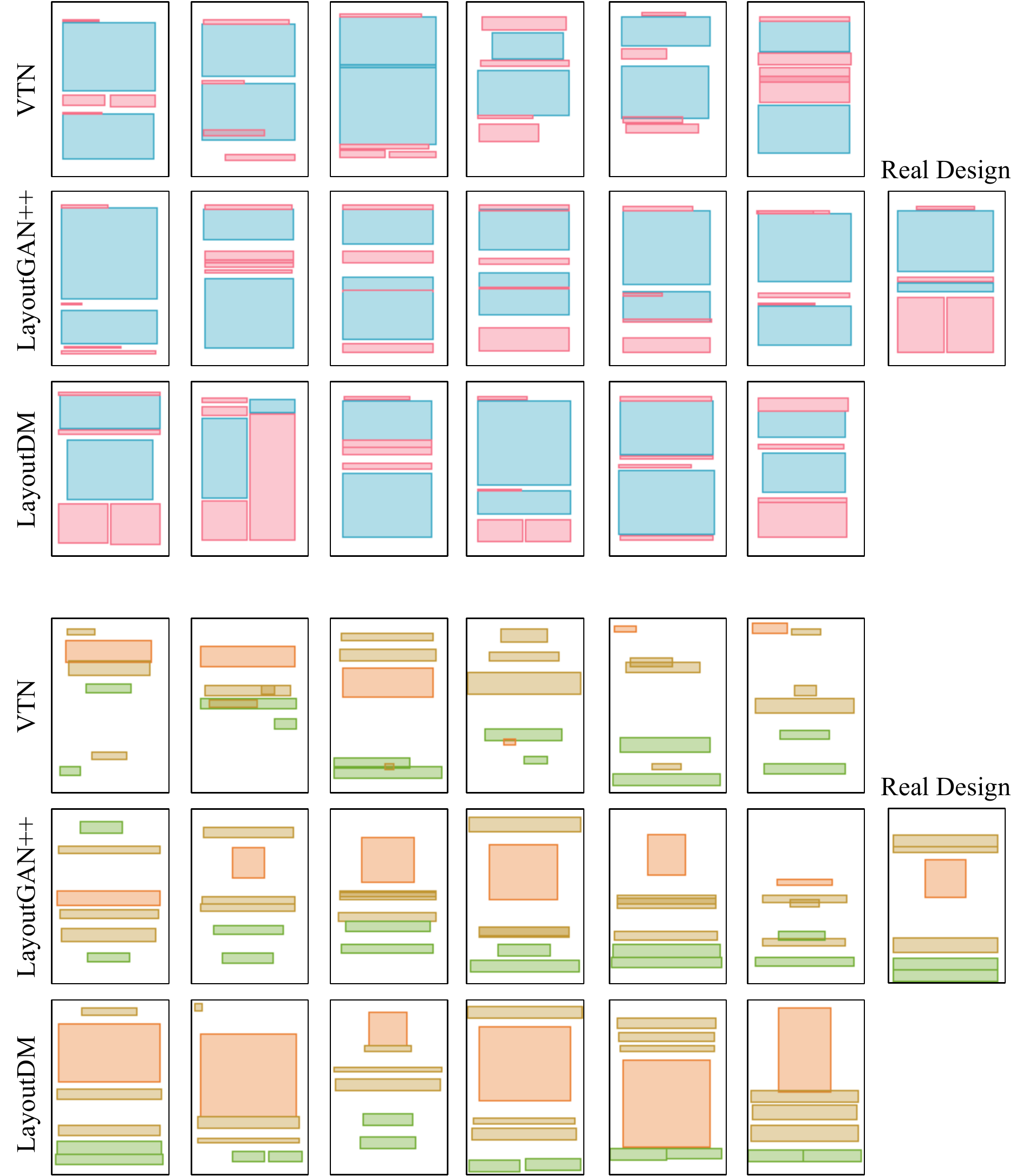}
    \caption{Additional quantitative diversity comparisons. The first to third rows are from the Publaynet dataset, and the fourth to sixth rows are from the Rico dataset. Layouts are generated conditioned on the same layout attributes (label categories) every three rows. Real designs are displayed for reference.}
    \label{fig:diversity_supplement}
\end{figure*}

We provide more quantitative comparisons on diversity in \cref{fig:diversity_supplement}. It can be observed that compared to the other two methods, our model has better diversity while maintaining high quality.

{\small
\bibliographystyle{ieee_fullname}
\bibliography{egbib}
}